# Uncertainty Estimation in Machine Learning

Valentin Arkov
*Automated Control and Management Dept.*
*Ufa State Aviation Technical University*
Ufa, Russia
ORCID: 0000-0002-7913-4778

*Abstract* — Most machine learning techniques are based upon statistical learning theory, often simplified for the sake of computing speed. This paper is focused on the uncertainty aspect of mathematical modeling in machine learning. Regression analysis is chosen to further investigate the evaluation aspect of uncertainty in model coefficients and, more importantly, in the output feature value predictions. A survey demonstrates major stages in the conventional least squares approach to the creation of the regression model, along with its uncertainty estimation. On the other hand, it is shown that in machine learning the model complexity and severe nonlinearity become serious obstacles to uncertainty evaluation. Furthermore, the process of machine model training demands high computing power, not available at the level of personal computers. This is why so-called pre-trained models are widely used in such areas of machine learning as natural language processing. The latest example of a pre-trained model is the Generative Pre-trained Transformer 3 with hundreds of billions of parameters and a half-terabyte training dataset. Similarly, mathematical models built from real data are growing in complexity which is accompanied by the growing amount of training data. However, when machine models and their predictions are used in decision-making, one needs to estimate uncertainty and evaluate accompanying risks. This problem could be resolved with non-parametric techniques at the expense of greater demand for computing power, which can be offered by modern supercomputers available, including those utilizing graphical and tensor processing units along with the conventional central processors.

## I. Introduction

In recent decades, machine learning techniques have become popular due to the huge amount of digital data available, enough computing power achieved, and the ease of use of free software libraries, such as Sci-Kit Learn for Python.

Machine learning is often defined as the branch of artificial intelligence dealing with extracting information from data. Computer algorithms can tune the model structure and parameters using sample training data. During the model training, the chosen quality criterion is optimized. The next step consists of model validation. This validation step requires another portion of the data, preferably not used in the model training. The validation might be done repeatedly with several data splits into training and testing sets. Therefore, it is usually called cross-validation. In this way, the model generalization ability is accounted for.

In many cases, machine learning techniques are inherited from well-established statistical routines, such as least squares (LS). Note that the LS procedure yields the model parameters estimates along with their uncertainty, typically in the form of the probability distribution. This distribution is characterized by the mean and variance. This analytical approach is based on the proposition that two parameters are sufficient in describing the distribution close to Gaussian, which is often the case for the LS method. This is then followed by the estimation of the confidence bands (intervals) given the confidence probability, derived from analytical techniques.

However, the existing ML tools are mostly focused on model training and validation, with much less attention paid to the analysis of statistical properties of the estimates obtained. Furthermore, cross-validation seems much less established, as compared to the model training process. A possible reason might be the limiter computing power available to the researchers. This is affirmed by that the LS criterion is still in wide use despite all its disadvantages. The LS technique was developed for manual computations, but it is still in use as the major quality criterion even in deep learning algorithms realized at the most powerful supercomputers.

In this paper, the uncertainty of the estimates obtained through model training is investigated using cross-validation techniques. The proposed approach is close to the simulation of uncertainty in metrology, rather than to the analytical ones used in statistical routines.

Data Science also includes statistical methods and techniques with applications in various areas such as business process automation, as described by Gonchar in [1].

## II. Quadratic Criteria in Machine Learning

Cross-validation appears to be a very compute-consuming technique. Thorough validation might take much greater computing power than the model training itself [2]. As a result, during cross-validation, only a very few splits are typically used. For example, Lakshminarayanan, Pritzel, and Blundell describe their experimentation with deep learning ensembles including uncertainty evaluation in [3]. In these experiments, multi-layered artificial neural networks were trained on standard ML data sets such as Boston housing. The authors mention that the chosen datasets were split into several train-test folds, with a maximum of 20 splits. The regression and classification accuracy and confidence are evaluated after the ML model has been trained. While some sort of uncertainty band for each metric is provided, the amount of cross-validation cannot be sufficient.

To emphasize the importance of evaluating uncertainty, the following four types of data analytics should be mentioned here. First, descriptive analytics corresponds to the branch of statistics dealing with various forms of distributions. Here we have a mere description of the past events, mostly presented in the form of means and variances.

Second, diagnostic analytics indicates possible causes for past events.

Third, predictive analytics is the tool for making predictions or forecasting. This implementation of machine models started in statistical learning in the form of regression models, see for example [4]. The authors emphasize that the incorporation of a vast number of features not relevant to the object under investigation does not necessarily improve the prognostic ability of the machine model, such as regression.

Prognostic systems can also include the evaluation of uncertainty, as indicated by Saxena in [5]. In that paper, uncertainty is presented as non-Gaussian non-parametric distribution to be approximated with some analytical function.

Fourth, prescriptive analytics usually provides suggested decisions to leverage the predictive power of the machine model, see for example papers by Vater, Harscheidt, and Knoll [6]. This stage is close to the process of the manufacturing management operation, rather than to mere analysis. Note that decision-making should include risk evaluation and management, which requires the estimation of the uncertainty for the predictions utilized.

The least squares technique is employed in modeling of both technical and economic systems, as exemplified by Wang in [7]. Mathematical modeling of technical objects is referred to as systems identification, while economic modeling is typically performed in the framework of econometric methodology. Despite the different terminology, one can easily reveal close similarities in the model representation.

Regression analysis in both systems identification and econometrics yields a similar model representation of coefficients along with their standard deviations, as shown below. This is then followed by the construction of confidence bands (limits) and significance testing under the chosen significance level. Again, this form of modeling is performed under the assumption of the normal distribution of the regression results, which is the implied consequence of the least-squares calculations.

Note that, unlike economic systems, technical ones are often manufactured at mass-production plants. Therefore, their construction and parameters are standardized and regulated by the design and technology documentation. This makes it possible to build precise mathematical models that only need to be refined individually using experimental data. This approach is described in detail, for example, by El-Sayed in [8].

Data mining represents another approach to searching for hidden patterns and interconnections within large data sets, see for example Leskovec [9]. Originally, this methodology was called Exploratory Data Analysis, compare for example the books by Tukey [10] and Bruce [11]. The techniques used for data mining and exploratory analysis still include regression with least-squares estimators of various kinds.

### III. MODEL UNCERTAINTY

Mathematical models obtained using machine learning methods represent more complex entities than conventional constructs. The number of model coefficients is stably increasing every year with more computing power available. For example, the Generative Pre-trained Transformer GPT-3 model based on deep learning uses hundreds of billions of parameters for natural language processing. This eliminates any attempt to analytically evaluate the possible uncertainty of each coefficient. GPT tools are often described in terms of possible application areas, with much less attention paid to their uncertainty, see for example a paper by Korngiebel and Mooney [12] discussing how the generative transformers are replacing live human interaction.

On the other hand, cross-validation techniques provide additional tools for numeric evaluation of uncertainty, while imposing growing requirements for an even greater amount of computation than for estimating the model itself.

The least-squares method for regression analysis is considered a well-established and researched approach, see for example works of Snedecor [13]. This type of analysis, generally supposes homogeneous variance of the random error, referred to as homoskedasticity in econometrics.

The fundamentals of the least squares for linear regression analysis are detailed in the literature on econometric methods, such as in books by Fomby, Johnson, and Hill [14]. This embraces necessary assumptions, and the Gauss-Markov theorem followed by a discussion of the concepts of convergence and consistency. The Gauss-Markov theorem reflects a generalized approach to regression analysis, as in the works by Kong [15], Lyche [16], Drygas [17], or Zimmerman [18].

Ideally, the estimates of the regression equation coefficients are accompanied by their standard deviations shown below the coefficients as follows:

$$y = a + b x \atop (\sigma_a) \quad (\sigma_b) \qquad (1)$$

Typically, the modeling pipeline includes the following stages:

1) Estimating the regression equation

2) Predicting point and interval values for the output feature

3) Plotting prediction/confidence intervals for the model output

4) Estimating residuals and analyzing the residuals plots

The last stage is connected to testing the conditions described in the Gauss–Markov Theorem, thus ensuring the desired properties of the estimates, such as asymptotic consistency, normality, and efficiency.

Consider a conceptual experiment described by Greene [19]. The proposed Monte Carlo study yields the non-parametric sampling distribution for the LS estimator. Slightly modify the experimental setup for our purposes as follows. The $x$ feature is drawn from the uniformly distributed population of random numbers. The $y$ feature is linearly coupled with $x$. We then add a normally distributed random disturbance $e$, see below.

$$\begin{aligned} y &= x - 100 + e \\ x &\sim U(150; 200) \\ e &\sim N(0; 10) \end{aligned} \qquad (2)$$

This training set is generated 1000 times with a sample size of 100 observations.

For every training set, the linear regression coefficients are estimated. Take the slope coefficient for further detailed examination. Using this "sample" of estimates for the model slope, their distribution is evaluated in the form of a histogram and a box-and-whiskers plot, see Figure 1. As the estimates' distribution is expected to be normal, the Gaussian curve is also created, with the same mean value and standard deviation as for the slope estimates sample.

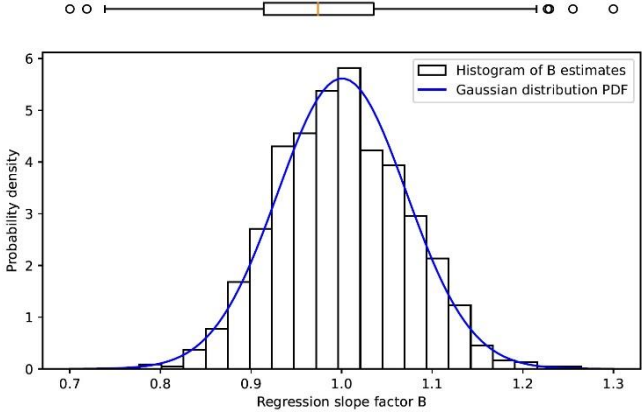

Fig. 1. Distribution of slope coefficient estimates

We can further expand the presentation of the simulation results. In Figure 2, the whole set of linear predictions is shown, along with the background filled with the training samples. One can observe that the prediction variance grows as the $x$ factor approaches the boundary of the known values. As we generate 1000 samples followed by estimating a linear model, the duration of the whole computational experiment is 1000 times greater than that of obtaining a single LS estimate.

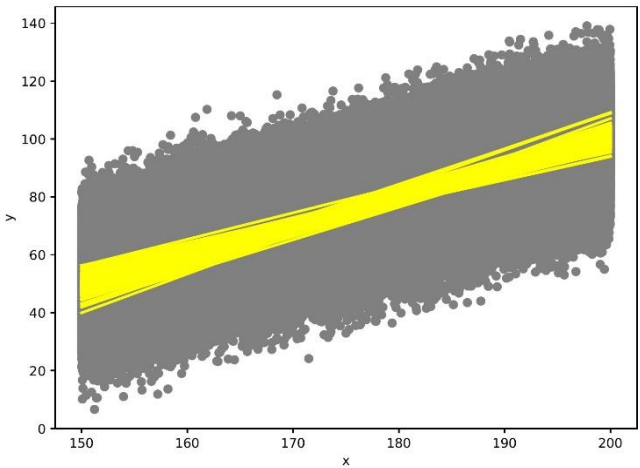

Fig. 2. Linear predictions over a series of samples

Following the analysis of variance for the estimates (usually referred to as ANOVA), the confidence bands are built as prediction intervals based on the prediction variance [20]. Thus the prediction uncertainty is evaluated in an "almost analytical" way, which is not feasible for complex ML models.

Extrapolating this analytical logic to machine learning, one could expect some sort of uncertainty estimation for the machine model parameters and predictions, as shown in Figure 3. The estimate of every parameter could be accompanied by its deviation, leading to the variation of the machine predictions. In most cases, however, this analytical approach is not feasible for several reasons, including the model nonlinearity. Therefore, LS techniques are focused on linear equations with nonlinear terms.

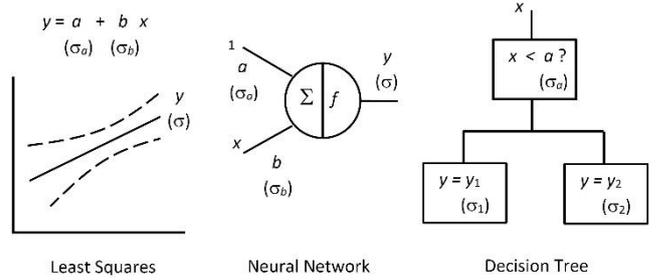

Fig. 3. Analytical estimation of uncertainty in machine learning

After the analysis of this historical background, one can discover that the machine learning routines very often possess much less depth in presentation and understanding. One might suppose that the deep analysis stage is often missing because the machine model/algorithm learns exclusively from the dataset and incorporates no prior/expert knowledge [21].

## IV. PROBLEM STATEMENT

The problem of estimating the uncertainty is formulated as follows. Given the training set of linearly correlated data, the uncertainty of the predicted output feature values is to be estimated in the form of the confidence interval, preferably non-sensitive to outliers. To demonstrate the proposed approach, a simplified problem statement is used, with the potential of further generalization to higher-order multi-dimensional modeling.

Consider the problem statement in more detail.

The source data for machine learning are presented in the ready-to-use tabular form, where the input and output features are organized in columns as ($x$, $y$). Suppose y is linearly correlated with $x$. The coefficients of the linear equation are unknown, with the presence of some additive random noise that is normally distributed.

An arbitrary machine learning model $M(c)$ is to be trained on the dataset available, with several parameters/coefficients c. The model training is performed to minimize the quadratic criterion comparing the actual and predicted output feature values: $\Sigma\Delta y^2 \to \min$. During training, the model structure and coefficients are chosen to generalize the data inter-relation through data train/test split in some proportion.

Given the input feature values within the known interval $X$, the output feature prediction $Y$ is then obtained using the trained model with the estimated model parameters $C$.

Finally, the uncertainty of the output predictions is to be estimated in the form of some interval, corresponding to the confidence band and less sensitive to possible outliers and fat tails of the distribution of **Y**. For demonstration purposes, two types of ML models are considered, namely the linear regression model and the random forest regressor.

## V. COMPUTATIONAL EXPERIMENT

Note that regression analysis exemplifies the use of numerical/non-parametric methods for data analysis, as described by Kiusalaas in [20]. The problem of estimating the uncertainty of regressor predictions is proposed to resolve with a non-parametric approach, similar to the Monte Carlo experiments described by Green in [19]. This approach is further extended to output value predictions as follows.

The training set is generated every time with a new random generator state, which is refreshed automatically with every function call. This is then followed by ML model training. For demonstration purposes, two extreme cases of ML models were chosen, namely linear regression and random forest regressor. Both models are based on the LS criterion in their training. The prediction factors are linearly organized with the **linspace** function to cover the interval of values in the training data.

During multiple runs of data generating and model training, predicted outputs are calculated for the same predictor factor values with every trained ML model. Eventually, the predicted output values are gathered into an array which is then processed to estimate the distribution of the predictions.

The general shape of the distribution is presented in Figure 4 as histograms for two predictions: *y*(150) and *y*(200). While the LS model coefficients and predictions have Gaussian distributions, the random forest regressor produces slightly different results. In order to indicate the difference, the ideal bell-shaped curves of the Gaussian probability density functions are also shown, with the actual values of the mean and standard deviation calculated for the array of the output predictions.

In this figure, box-and-whiskers plots are also shown for these two predicted output values. As one can easily see, both distributions demonstrate much longer tails with a lot of outliers outside the standard quartile deviation borders. The central part of both histograms is narrow as compared to the Gaussian curve.

Figure 5 shows the total array of the training sets (grey markers) along with the regressor predictions. The central tendency corresponding to the regression line is obtained as the median of all output predictions. The lower and upper bands for the predictions are calculated as the quartile deviation from the median in the way used in box plots as follows.

The quartiles Q1, Q2, and Q3 are obtained for the predictions using interpolation where appropriate. The Inter-Quartile Range (IQR) indicated the distance between the third and first quartiles. The lower and upper limits are then obtained to approximate the "three sigma" confidence interval for the confidence probability of about 99.7%.

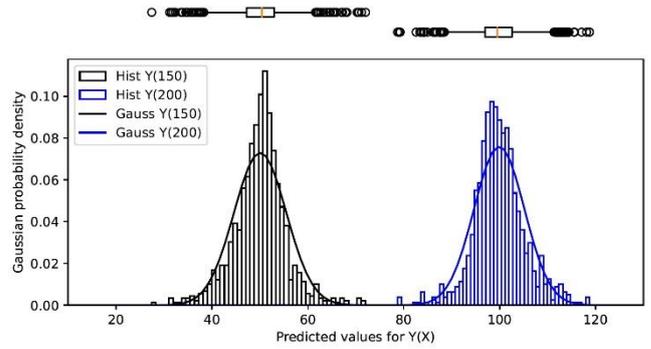

Fig. 4. Boxplots and histograms for random forest predictions

The upper and lower limits/bands for the predicted values are built as follows.

$$
\begin{aligned}
&\text{IRQ} = Q3 - Q1 \\
&\text{Median} = Q2 \\
&\text{Low limit} = Q1 - 1.5 \text{ IQR} \\
&\text{Upper limit} = Q3 + 1.5 \text{ IQR}
\end{aligned}
\quad (3)
$$

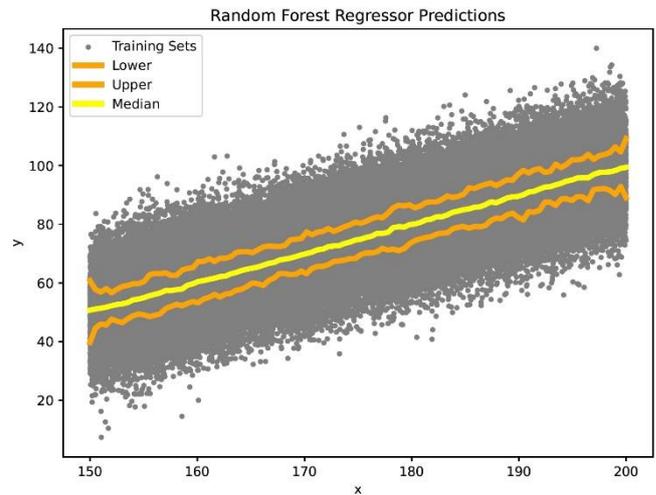

Fig. 5. Uncertainty of random forest prediction shown as quartile deviation

In addition to the estimation of uncertainty, one can discover the smoothing effect of the utilization of quartile measures. Both median and limit bands look quite smooth, as opposite to the predictions obtained on a single sample, see Figure 6.

In both Figures 5 and 6, the model prediction follows the general pattern of the least-squares estimate which produces a slightly lesser slope of the prediction line. This means that the slope regression coefficient is less than the original value in the model used in the data generation.

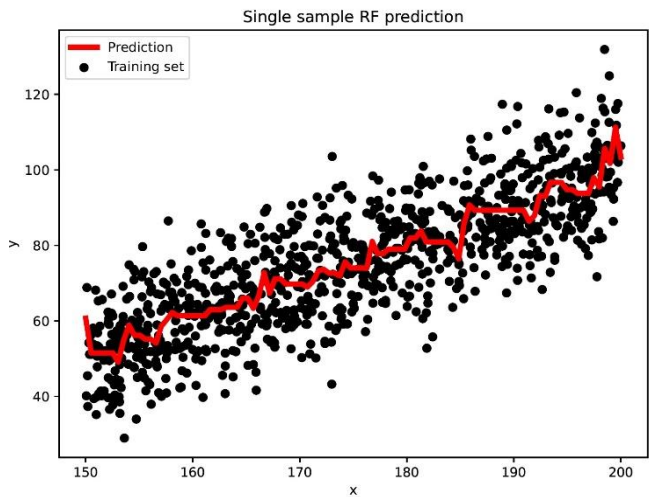

Fig. 6. Output predictions of random forest regressor for single training sample

Note that similar behavior is observed in most machine learning models using LS criteria, usually followed by the model quality evaluation with the mean squared error (MSE).

Recent developments in interactive parallel computations offer more possibilities for machine learning applications on parallel apps like R or Matlab [22]. In particular, interactive machine learning becomes available for Python programs utilizing TensorFlow [23]. This is further added with the automatic machine learning (AutoML) tools, thus making data analysis and machine learning easy to use for wider numbers of researchers. A popular example for AutoML would be Auto-sklearn [24], with much attention paid to hyperparameter optimization. Note that at each step of model training, Auto-sklearn produces an ensemble of models. This is followed by estimating uncertainty in the form of the median error with $5^{th}$ and $95^{th}$ percentiles, which is shown as a demonstration by-product.

## VI. CONCLUSIONS AND FUTURE WORK

In this work, the basics of regression analysis have been extrapolated to demonstrate the need for uncertainty estimation in machine learning, currently absent in modern software kits. The proposed non-parametric approach is based on cross-validation and might yield uncertainty estimates for various types of machine learning predictions.

While the analytical estimation of uncertainty is not feasible because of the substantial model nonlinearity, the non-parametric approach can offer an alternative method for dealing with predictions. Furthermore, standard deviation as a measure of uncertainty is justified for Gaussian disturbances, which is not always the case in the real world. Instead, box-and-whiskers plots and quantile measures for uncertainty present a much more flexible and promising tool.

Further developments in the proposed uncertainty estimation approach might embrace other ML techniques, such as classification, clustering, and dimensionality reduction. When dealing with the actual big data, one would need to use supercomputers, as the proposed non-parametric estimation of uncertainty requires much greater computing power. The available graphical processing units could offer the acceptable level of computation speedup to perform multiple runs of model training and cross-validation resulting in quality estimates.